\pdfoutput=1
\documentclass[11pt]{article}

\usepackage[final]{acl}

\usepackage{times}
\usepackage{latexsym}
\usepackage[T1]{fontenc}
\usepackage[utf8]{inputenc}
\usepackage{microtype}
\usepackage{inconsolata}
\usepackage{graphicx}
\usepackage{amsmath}
\usepackage{amssymb}
\usepackage{booktabs}
\usepackage{multirow}
\usepackage[safe]{tipa}

\title{Phonologically Informed Tokenization for German Speech Recognition: \\
A Cross-Domain Study}

\author{Christopher Witzl \quad Tobias Bocklet \quad Korbinian Riedhammer \\
        Technische Hochschule Nürnberg Georg Simon Ohm, Nürnberg, Germany \\
        \texttt{\{christopher.witzl,\,tobias.bocklet,\,korbinian.riedhammer\}@th-nuernberg.de}}

\begin{document}
\maketitle

\begin{abstract}
German is a morphologically rich language whose syllable structure is exceptionally well-predicted by the Knuth--Liang hyphenation algorithm.
We ask whether phonologically informed tokenization can serve as a competitive target for end-to-end speech recognition.
We compare three tokenizer families on the Omnilingual ASR wav2vec 2.0 backbone fine-tuned with CTC: the pretrained multilingual character inventory, a data-driven Byte-Pair Encoding (BPE) over orthography, and phonologically informed units from Pyphen syllabification and grapheme-to-phoneme conversion.
Across 40 fine-tunes, we evaluate on three German test sets spanning orthogonal shifts: in-domain read speech, dialectal spontaneous speech, and standard-German spontaneous speech.
In-domain, all phonologically informed tokenizers match BPE and the multilingual character baseline on both WER and CER.
Under domain shift the picture splits along vocabulary size rather than the linguistic axis of variation: at small vocabularies, syllable-aware tokenization improves on dialectal speech, where phonetic surface forms vary but syllable structure is preserved, and stays ahead on spontaneous speech, where new word-forms violate vocabulary closure.
A phoneme-level confusion analysis further shows that all tokenizers commit the same canonical function-word errors, indicating that the acoustic encoder, not the tokenizer, dominates the error topology.
Our findings suggest that tokenizer choice may depend on the vocabulary budget as much as on the distribution shift expected at deployment rather than reducing to a single universal optimum.
\end{abstract}

\section{Introduction}
\label{sec:intro}

Modern end-to-end automatic speech recognition (ASR) systems built on self-supervised pretraining \citep{baevski2020wav2vec2} couple an acoustic encoder with an output tokenizer that determines the units the network must predict.
The de facto default has converged on two choices: a data-driven Byte-Pair Encoding (BPE) \citep{sennrich2016bpe} fitted to the training transcripts, or a fixed multilingual character inventory inherited from the pretrained ASR model \citep{omnilingualasr2025}.
Both are phonologically agnostic.
The first treats character $n$-grams as merge candidates without regard to syllable or morpheme structure; the second models every character of every supported script as an independent unit, leaving word-level structure to the acoustic encoder.
The choice is not cosmetic: the tokenizer fixes which strings the network can emit, how many target units per word the CTC objective must align, and how the model degrades on word-forms unseen in training.

This convergence is convenient, but it hides a methodological question with real downstream consequences: do \emph{phonologically informed} tokenizers\footnote{I.e., tokenizers whose boundaries derive from linguistic knowledge (syllabification, grapheme-to-phoneme mappings) rather than data-driven frequency statistics.},
those that align with the syllabic and phonemic structure of the target language, offer something that data-driven and multilingual defaults cannot?
German is a natural test case: it is morphologically rich (productive compounding, inflection), it has substantial dialectal variation, and large-scale German ASR benchmarks (Common Voice~\citep{ardila2020commonvoice}, Spoken Wikipedia Corpus~\citep{kohn2016swc}) coexist with structurally different out-of-distribution (OOD) corpora, namely dialectal spontaneous speech (BAS RVG\,1, \citealp{burger1998rvg1}) and standard-German spontaneous dialogues (Verbmobil, \citealp{burger2000verbmobil}).
The language thus offers two distinct OOD axes against which tokenizer behavior can be probed: RVG\,1 varies the \emph{phonetic surface} (a largely shared lexicon realized across nine dialect regions), while Verbmobil varies the \emph{lexical inventory} (standard-German phonetics but negotiation-dialogue vocabulary including hesitations, discourse particles, and colloquial verb forms unseen at training time).

\paragraph{Contributions.}
We make three contributions.
(1) We assemble and train \textbf{40 CTC fine-tunes of the Omnilingual~ASR wav2vec\,2.0 encoder} \citep{omnilingualasr2025} (31 tokenizers at $300\,$M, nine at $1\,$B) that share encoder, training data, and training schedule, varying only the tokenizer: 13 orthographic tokenizers (character, BPE-512/1024/2048, syllable-512/1024/2048, multi-syllable with and without syllable separators) and 18 phoneme-track tokenizers (raw IPA, phoneme-BPE, phoneme-syllable, phoneme-multi-syllable; vocabulary sizes $43$--$3{,}840$).
The matrix is large enough to dissociate vocabulary size from tokenization strategy.
(2) We evaluate every model on the in-domain test set \emph{and} on two structurally orthogonal OOD test sets: dialectal spontaneous speech (RVG\,1) and standard-German negotiation dialogues (Verbmobil), quantifying \emph{when} the phonological bias helps, \emph{where} it stays competitive, and \emph{how} its payoff depends on vocabulary size.
(3) We complement aggregate WER with paired-bootstrap significance testing \citep{bisani2004bootstrap} and confusion-matrix analyses; we focus on WER and CER rather than token-level metrics, since the latter conflate recognition quality with tokenizer compression rate and are not directly comparable across the very different vocabulary sizes (43 to 10,288 tokens) in our design space.

\paragraph{Headline result.}
Phonologically informed tokenizers cluster within the confidence intervals of BPE and character baselines in-domain.
Under dialectal-phonetic OOD they show a small advantage that sits at the Bonferroni boundary ($-1.09$\,pp WER on RVG\,1-ORT, $p{=}0.019$, threshold $\alpha/k{\approx}0.017$ for $k{=}3$ tests on this dataset), and under spontaneous-lexical OOD a small lead ($-0.89$\,pp WER for syllable-512 on Verbmobil), giving a characterization of the phonological-bias trade-off rather than a single ``best tokenizer''.
That trade-off runs along vocabulary size rather than along the type of shift: the advantage reverses on both corpora at $|V|{=}2048$ (§\ref{sec:vocab}).

\section{Related Work}
\label{sec:related}

The choice of output unit in ASR has recurred as a design question through every architectural paradigm.
Classical large-vocabulary systems combined pronunciation lexica, context-dependent triphones, and language models \citep{lee1990sphinx,young1994treebased}, later composed through weighted finite-state transducers \citep{mohri2002wfst}.
For German specifically, \citet{killer2003grapheme} found that graphemes match phonemes in GMM-HMM systems, while \citet{mousa2012morpheme} reported that morpheme-level language models lower out-of-vocabulary rates. Both findings are early evidence that linguistic units remain competitive in-domain, with conditional advantages elsewhere.

The unit question survived the neural transition.
Hybrid NN/HMM systems \citep{bourlard1994connectionist} retained phonemic states as targets; end-to-end CTC reopened the question, with \citet{graves2013speech} reestablishing phoneme-CTC on TIMIT.
\citet{sennrich2016bpe} introduced Byte-Pair Encoding for neural machine translation, and \citet{zenkel2018subword} adapted it to CTC ASR with a key contrast between \emph{Subword} units (merges constrained to within words) and \emph{Crossword} units (merges allowed across word boundaries).
Their distinction maps directly onto our \emph{syllable} vs.\ \emph{multisyl} contrast at the syllable rather than word level.

\citet{irie2019choice} compared graphemes, phonemes, and wordpieces on LibriSpeech and found graphemes and wordpieces dominate without an external language model; for German, \citet{khosravani2020comparison} found BPE and unigram subwords to outperform character- and word-level outputs (with $n$-gram LMs).
\citet{sainath2018nolexicon} found that graphemes largely obviate the lexicon, whereas \citet{zeineldeen2020systematic} responded that phoneme-BPE with homophone disambiguation remains competitive.
A separate line of work derives subword vocabularies from pronunciation information \citep{xu2019pasm,papadourakis2021phis,zhou2021adsm}.
Most directly related to our phoneme track, \citet{klumpp2022commonphone} released the Common Phone corpus and trained a wav2vec\,2.0 phoneme recognizer covering six languages including German, achieving $18.1$\,\% PER across $101$ phonetic symbols; our phoneme-track models follow the same acoustic-modeling setup but vary the tokenizer.

Syllable-based ASR has a thinner literature.
\citet{ganapathiraju2001syllable} demonstrated syllabic acoustic units in HMM-LVCSR on Switchboard; \citet{qu2017syllable} trained syllable-level CTC for Mandarin; and \citet{manohar2023sbpe} applied Syllable-BPE to Malayalam, the closest cross-linguistic precedent to our German Pyphen-based syllable tokenizers \citep{liang1983hyphenation,pyphen}.
None of these works evaluated tokenizers under structured domain shift, which is the gap our paper addresses.

A complementary literature on ASR robustness holds the tokenizer fixed and varies the acoustic side.
\citet{likhomanenko2021rethinking} established that single-corpus WER is a poor robustness proxy; \citet{hsu2021robust} showed that encoder-side adaptation closes $66$--$73$\,\% of the OOD gap in wav2vec\,2.0; and \citet{radford2023whisper} framed robustness through weak-supervision scale.
Our results are consistent with the encoder-as-primary-lever finding (confusion topology is largely tokenizer-invariant in our setup), but tokenizer choice acts as a complementary lever whose payoff depends on the OOD axis and vocabulary size.
We extend this lineage by holding the encoder fixed and varying only the output tokenizer, isolating the tokenization effect from the lexicon, decoder, and training-scale variables that confounded earlier comparisons.

\section{Methodology}
\label{sec:method}

\subsection{Datasets}
\label{sec:data}

Training data is a $961$-hour mixture of German Common Voice \citep{ardila2020commonvoice} and the Spoken Wikipedia Corpus \citep[Oct.\ 2017 release;][]{kohn2016swc,baumann2019swc}.
Evaluation uses three structurally distinct test sets: the held-out portion of the same mixture (CV+SWC test, in-domain), BAS RVG\,1 \citep{burger1998rvg1}, and BAS Verbmobil~VM1 \citep{burger2000verbmobil,schiel1999bas}.
Table~\ref{tab:corpora} summarizes the training mixture and the three evaluation sets.

\begin{table*}[t]
  \centering
  \small
  \setlength{\tabcolsep}{6pt}
  \begin{tabular}{llrrl}
    \toprule
    \textbf{Corpus} & \textbf{Split} & \textbf{$n$ utterances} & \textbf{hours} & \textbf{Role / OOD axis} \\
    \midrule
    CV~+~SWC                                    & train           & 550\,747 & 960.5 & training mix \\
    \quad CV~24~DE \citep{ardila2020commonvoice} & (part of train) & 457\,512 & 726.5 & speakers worldwide \\
    \quad SWC DE \citep{kohn2016swc}                & (part of train) &  93\,235 & 234.0 & Spoken Wikipedia \\
    \midrule
    CV~+~SWC                                    & test            &  24\,604 &  50.2 & in-domain hold-out \\
    RVG\,1 \citep{burger1998rvg1}               & all, pooled     &      442 &   6.6 & dialectal-phonetic OOD \\
    Verbmobil~VM1 \citep{burger2000verbmobil}   & all, pooled (German) &  13\,633 &  33.8 & spontaneous-lexical OOD \\
    \bottomrule
  \end{tabular}
  \caption{Corpora. Training uses the CV\,24~+~SWC mixture; the in-domain
  test split is held out from the same mixture; RVG\,1 and Verbmobil are
  entirely unseen and pooled across their train/dev/test splits
  (§\ref{sec:eval}). Both OOD corpora carry parallel ORT (canonical
  orthography) and TR2 (dialect-/prosody-annotated transliteration) tiers.
  Verbmobil~VM1 also contains English negotiation dialogues; we evaluate on
  its German subset only (§\ref{sec:eval}).
  \label{tab:corpora}}
\end{table*}

The two OOD corpora are intentionally orthogonal.
\textbf{RVG\,1} is spontaneous speech recorded across nine German dialect regions; it varies the phonetic realization while the underlying lexicon remains largely shared across regions.
\textbf{Verbmobil~VM1} is spontaneous standard-German negotiation speech; it keeps the phonetic norm but introduces unfamiliar word-forms (hesitations, repairs, colloquial shortenings, discourse particles).
Each corpus therefore stresses a different tokenizer property: phonetic robustness for RVG\,1, lexical coverage for Verbmobil.
Both provide parallel ORT (canonical orthography) and TR2 (dialect transliteration with prosody markers) reference tiers, which we report separately.

\subsection{Tokenizer design space}
\label{sec:tokenizer-space}

We compare three classes of tokenizers, deliberately matched for vocabulary size where possible.
Table~\ref{tab:tokenizers} shows what the smallest variant of each family produces on a single illustrative example, which we discuss in §\ref{sec:tokenizer-comparison}.

\subsection{Multilingual character inventory}
\label{sec:omni-vocab}

\textsc{Omni-vocab} is the publicly released \texttt{omniASR\_tokenizer\_written\_v2} of Omnilingual~ASR \citep{omnilingualasr2025}, declared as \texttt{tokenizer\_family:\,char\_tokenizer} in the model card: a character-level tokenizer with $\approx 10{,}288$ multilingual code-points covering 1{,}600+ languages, shipped as the default for all v2 architectures (CTC and LLM checkpoints from $300\,$M to $7\,$B).
On our German training data, only $59$ of these code-points occur more than $100$ times each (top-$50$ coverage $>$$99.99$\,\% of all token occurrences); a long tail of $\sim$$600$ further code-points appears as rare orthographic artifacts.
\textsc{Omni-vocab} serves as the ``no phonological bias'' comparator inherited from the pretrained ASR model.

\subsection{Data-driven subword}
\label{sec:german-bpe}

\textsc{German-BPE} is a standard SentencePiece BPE \citep{sennrich2016bpe,kudo2018sentencepiece} fitted to the German training transcripts at vocabularies $\{512, 1024, 2048\}$.
It serves as the data-driven comparator and is the de facto default in many modern German ASR systems.

\subsection{Phonologically informed orthographic tokenizers}
\label{sec:phon-ortho}

We construct three orthographic families on top of SentencePiece BPE \citep{sennrich2016bpe,kudo2018sentencepiece}, differing only in which symbols are protected as user-defined symbols (UDS) and whether the training text is pre-syllabified.
\textsc{Syllable-$k$} uses the Knuth--Liang hyphenation algorithm \citep{liang1983hyphenation} via Pyphen \citep{pyphen} to syllabify the training transcripts; the syllable separator \texttt{-} is registered as a UDS so that no BPE merge crosses a syllable boundary.
Vocabularies are $k \in \{512, 1024, 2048\}$.
\textsc{Multisyl-$k$} is trained on the same syllabified text, but the separator \texttt{-} is \emph{not} protected, allowing merges to cross syllable boundaries; in addition, the $100$ most frequent German syllables are pre-registered as UDS so the model retains the most useful multi-character atoms even at small vocabulary sizes.
This corresponds conceptually to \citeposs{zenkel2018subword} \emph{Crossword} variant but operates at the syllable rather than word level.
\textsc{Multisyl-nosep-$k$} keeps the same $100$-syllable UDS list as Multisyl but is trained on \emph{raw}, non-syllabified text, so no syllable-separator token ever appears in the inputs.
It isolates the effect of having syllables as atomic vocabulary entries from the effect of an explicit syllable-boundary marker.

\subsection{Phonologically informed phoneme tokenizers}
\label{sec:phon-phone}

The phoneme track applies the same four merge strategies (raw, BPE, syllable, multi-syllable) to G2P-derived X-SAMPA phoneme sequences \citep{bisani2008joint}.
The G2P model is a joint-sequence WFST built with Phonetisaurus \citep{novak2012phonetisaurus} and trained on the German pronunciation lexicon shipped with MaryTTS \citep{schroder2003marytts} (\texttt{de.txt}, $\sim$$31$k entries), giving canonical SAMPA transcriptions for every word in our training and test transcripts.
Out-of-lexicon words are handled by Phonetisaurus' default joint-sequence backoff (beam~size~$20$); compound forms are not pre-decomposed and are mapped as whole tokens.
The 18 resulting phoneme-token configurations span vocabulary sizes from $43$ (the smallest character-level raw variant) to $3{,}840$ (the largest multi-syllable BPE).
The phoneme target inherits the German MAUS/BAS tradition \citep{schiel1999bas} and parallels Common Phone's wav2vec\,2.0 phoneme recognition setup \citep{klumpp2022commonphone}.

\subsection{Tokenizer comparison}
\label{sec:tokenizer-comparison}

Table~\ref{tab:tokenizers} compares the smallest variant of each tokenizer family on a single illustrative word, both in its orthographic form \emph{Unterhaltung} (12 characters, four canonical syllables \textbf{un-ter-hal-tung}) and in its X-SAMPA transcription \texttt{?Un-t6-hal-tUN}.
On the orthographic track, \emph{syllable-512} is the only family that recovers all four canonical syllables as individual tokens; \emph{German-BPE-512} finds three of them from data alone but splits the suffix \texttt{tung} into \texttt{t\,|\,ung}; the multisyl families pick finer, phonotactically motivated clusters that can additionally merge across syllable boundaries.
On the phoneme track, \emph{phone-multisyl-lite-512} is the only family that captures all four SAMPA syllables as single tokens.
The classes therefore differ in \emph{what kind of boundary} they choose, not just in granularity; this distinction frames the cross-domain results in §\ref{sec:results}.

\begin{table*}[t]
  \centering
  \footnotesize
  \setlength{\tabcolsep}{4pt}
  \begin{tabular}{llrrrl}
    \toprule
    \textbf{Family} & \textbf{Variant} & \textbf{Vocab} & \textbf{$n_\mathrm{tok}$} & \textbf{cpt} & \textbf{Tokenization} \\
    \midrule
    \multicolumn{6}{l}{\emph{Orthographic track:} segmenting the German word ``\texttt{unterhaltung}'' (12 chars, 4 canonical syllables un-ter-hal-tung)} \\
    \midrule
    Multilingual char         & omni-vocab                 & 10\,288 & 12          & 1.00          & \texttt{u\,|\,n\,|\,t\,|\,e\,|\,r\,|\,h\,|\,a\,|\,l\,|\,t\,|\,u\,|\,n\,|\,g} \\
    Data-driven BPE           & german-BPE-512             &    512  & 5           & 2.40          & \texttt{un\,|\,ter\,|\,hal\,|\,t\,|\,ung} \\
    \textbf{Syllable (Pyphen)} & \textbf{syllable-512}     &    512  & \textbf{4}  & \textbf{3.00} & \texttt{\textbf{un\,|\,ter\,|\,hal\,|\,tung}} \\
    Multisyl (cross-syllable) & multisyl-512               &    512  & 6           & 2.00          & \texttt{un\,|\,ter\,|\,ha\,|\,lt\,|\,un\,|\,g} \\
    Multisyl-nosep            & multisyl-nosep-512         &    512  & 6           & 2.00          & (identical to multisyl-512 on a single word$^{\dagger}$) \\
    \midrule
    \multicolumn{6}{l}{\emph{Phoneme track:} segmenting the X-SAMPA transcription ``\texttt{?Un-t6-hal-tUN}'' of the same word} \\
    \midrule
    Atomic phones             & phone-char-raw             &  43     & 14          & 0.79          & \texttt{?\,|\,U\,|\,n\,|\,-\,|\,t\,|\,6\,|\,-\,|\,h\,|\,a\,|\,l\,|\,-\,|\,t\,|\,U\,|\,N} \\
    Data-driven BPE           & phone-bpe-128              & 128     & 10          & 1.10          & \texttt{?\,|\,Un\,|\,-\,|\,t\,|\,6-\,|\,h\,|\,al\,|\,-\,|\,t\,|\,UN} \\
    Syllable                  & phone-syllable-128         & 128     & 11          & 1.00          & \texttt{?\,|\,Un\,|\,-\,|\,t\,|\,6\,|\,-\,|\,h\,|\,al\,|\,-\,|\,t\,|\,UN} \\
    \textbf{Multisyl-lite}    & \textbf{phone-multisyl-lite-512} & 512 & \textbf{7} & \textbf{1.57} & \texttt{\textbf{?Un\,|\,-\,|\,t6\,|\,-\,|\,hal\,|\,-\,|\,tUN}} \\
    \bottomrule
  \end{tabular}
  \caption{Smallest variant of each tokenizer family on the German word
  \emph{Unterhaltung} (orthographic track) and its X-SAMPA transcription
  \texttt{?Un-t6-hal-tUN} (phoneme track). Bold rows mark the family
  that recovers all four canonical syllables (or SAMPA syllables) as
  individual tokens. cpt = characters per token; on the phoneme track,
  atomic SAMPA symbols are counted, excluding syllable-separator
  dashes. $^{\dagger}$multisyl and multisyl-nosep produce identical
  outputs on a single word; the boundary marker affects only
  vocabulary construction.
  \label{tab:tokenizers}}
\end{table*}

\subsection{Evaluation protocol}
\label{sec:eval}

We report token-weighted \textbf{WER and CER} for orthographic models and \textbf{PER and CER} for phoneme-track models, both with $95$\,\% paired-bootstrap confidence intervals (CIs; $n_\mathrm{boot}{=}10{,}000$, seed=$42$, percentile method) \citep{bisani2004bootstrap}.
For the orthographic track, references and hypotheses are normalized before scoring (lowercase, ß~$\!\to\!$~ss, punctuation stripped, whitespace collapsed); this matches the standard Whisper-style protocol and ensures that capitalization conventions of the reference (Verbmobil business-style transcripts use Sie/Termin/Mai capitalization, while CV+SWC are lowercase) do not contaminate the recognition metric.
Phoneme-track (X-SAMPA) evaluation is not normalized, since SAMPA symbols would be destroyed by punctuation stripping.
Pairwise comparisons apply per-dataset Bonferroni correction.
For the OOD corpora we pool train+dev+test (none seen at training time, which justifies pooling for tighter intervals at no leakage cost).
Verbmobil~VM1 contains English as well as German negotiation dialogues; we score its German subset ($13{,}633$ utterances, $33.8$\,h), identified from the reference transcripts by a German/English function-word criterion.
Audio is $16\,$kHz mono and capped at $60\,$s for CV+SWC and Verbmobil (matching the external baselines); RVG\,1's full-length recordings, the only ones routinely exceeding $60\,$s, are instead segmented at silence boundaries into ${\le}60\,$s chunks so the complete utterance is scored rather than truncated.
Decoding is greedy CTC without an external language model: a token-level LM would tie the decoding vocabulary to the tokenizer under study and conflate its language-modeling contribution with the acoustic comparison.
Every WER/CER/PER reported here therefore reflects the encoder plus tokenizer-specific head in isolation; LM rescoring, and the tokenizer--LM interaction it opens up, is left to future work (§\ref{sec:conclusion}).

\section{Experiments and Results}
\label{sec:results}

\subsection{Setup}
\label{sec:experiments}

We train 40 CTC fine-tunes \citep{graves2006ctc} of the Omnilingual~ASR wav2vec\,2.0 backbone \citep{baevski2020wav2vec2,omnilingualasr2025}: all 31 tokenizers (13 orthographic, 18 phoneme-track) at $300\,$M, and a nine-tokenizer subset at $1\,$B. The subset comprises the full vocabulary sweep of Syllable and German-BPE plus one variant per remaining family (omni-vocab, multisyl-512, multisyl-nosep-512).
All runs use the Adam optimizer with a fixed learning rate of $5{\times}10^{-5}$, bfloat16 mixed precision, no encoder freezing, $20\,$k optimization steps, and a batched dynamic budget of $\approx$$2.3{\times}10^{7}$ audio elements per gradient step (min/max audio $2$/$60\,$s).
Batching is audio-element-based rather than token-based: every run sees the same audio in the same order (identical seed) for the same $20\,$k updates, and the batch size is never adjusted per tokenizer.
The number of target tokens therefore varies with tokenizer fertility; we deliberately do not match it, since fixing the token count would unfix the audio budget and confound the comparison with the amount of acoustic evidence.
Appendix~\ref{app:budget} reports the resulting target-token counts per run.
The CTC output head is a single linear projection from the final encoder layer to the tokenizer's vocabulary size (no layer selection was performed), with Xavier-uniform weight initialization and zero-initialized bias; no final-layer dropout is applied.
Augmentation, label smoothing, validation schedule, and early-stopping criterion are held identical across all runs so that any cross-tokenizer difference is attributable to the output unit and its head alone.
We organize the empirical comparison around three research questions, each probing one of our three test sets, complemented by paired significance tests for nine focal comparisons.

\subsection{Q1: In-domain WER and CER}

\begin{table}[t]
  \centering
  \footnotesize
  \setlength{\tabcolsep}{3pt}
  \begin{tabular}{lr@{\ }lr@{\ }l}
    \toprule
    \textbf{Tokenizer} & \textbf{WER\,\%} & \textbf{95\,\% CI} & \textbf{CER\,\%} & \textbf{95\,\% CI} \\
    \midrule
    \multicolumn{5}{l}{\emph{Multilingual character inventory}} \\
    omni-vocab           & 15.64          & {\scriptsize[15.38, 15.90]} & \textbf{6.09} & {\scriptsize[5.89, 6.29]} \\
    \midrule
    \multicolumn{5}{l}{\emph{Data-driven BPE (German)}} \\
    German-BPE-512       & \textbf{14.93} & {\scriptsize[14.66, 15.18]} & 6.14          & {\scriptsize[5.94, 6.35]} \\
    German-BPE-1024      & 16.88          & {\scriptsize[16.61, 17.13]} & 6.89          & {\scriptsize[6.69, 7.09]} \\
    German-BPE-2048      & 16.11          & {\scriptsize[15.84, 16.36]} & 6.65          & {\scriptsize[6.45, 6.85]} \\
    \midrule
    \multicolumn{5}{l}{\emph{Syllable (Pyphen)}} \\
    Syllable-512         & 15.16          & {\scriptsize[14.90, 15.42]} & 6.23          & {\scriptsize[6.02, 6.43]} \\
    Syllable-1024        & 15.44          & {\scriptsize[15.18, 15.69]} & 6.36          & {\scriptsize[6.15, 6.56]} \\
    Syllable-2048        & 15.94          & {\scriptsize[15.68, 16.20]} & 6.48          & {\scriptsize[6.28, 6.68]} \\
    \midrule
    \multicolumn{5}{l}{\emph{Multisyl (cross-syllable merges)}} \\
    Multisyl-512         & 15.72          & {\scriptsize[15.46, 15.98]} & 6.46          & {\scriptsize[6.25, 6.67]} \\
    Multisyl-1024        & 16.27          & {\scriptsize[16.00, 16.53]} & 6.63          & {\scriptsize[6.42, 6.83]} \\
    Multisyl-2048        & 15.72          & {\scriptsize[15.46, 15.98]} & 6.39          & {\scriptsize[6.18, 6.60]} \\
    \midrule
    \multicolumn{5}{l}{\emph{Multisyl-NoSep (raw input, no syllable marker)}} \\
    Multisyl-nosep-512   & 15.87          & {\scriptsize[15.61, 16.13]} & 6.42          & {\scriptsize[6.22, 6.63]} \\
    Multisyl-nosep-1024  & 16.37          & {\scriptsize[16.11, 16.62]} & 6.58          & {\scriptsize[6.37, 6.78]} \\
    Multisyl-nosep-2048  & 18.00          & {\scriptsize[17.73, 18.26]} & 6.99          & {\scriptsize[6.78, 7.20]} \\
    \bottomrule
  \end{tabular}
  \caption{In-domain WER and CER on CV+SWC test ($n{=}24{,}604$),
  $300\,$M wav2vec\,2.0 with CTC, on normalized text (lowercase,
  ß~$\!\to\!$~ss, punctuation stripped). Brackets: paired-bootstrap
  95\,\% CIs ($n_\mathrm{boot}{=}10{,}000$, seed=$42$); bold marks the
  per-metric best across the 13 orthographic tokenizers.
  \label{tab:in-domain}}
\end{table}

\begin{table}[t]
  \centering
  \footnotesize
  \setlength{\tabcolsep}{3pt}
  \begin{tabular}{lrr@{\ }l}
    \toprule
    \textbf{Phone Tokenizer} & \textbf{Vocab} & \textbf{PER\,\%} & \textbf{95\,\% CI} \\
    \midrule
    \multicolumn{4}{l}{\emph{Atomic / raw}} \\
    phone-char-raw-v2          & 43           & 18.05 & {\scriptsize[17.78, 18.33]} \\
    phone-raw-v2               & 50           & 18.35 & {\scriptsize[18.08, 18.64]} \\
    \midrule
    \multicolumn{4}{l}{\emph{Phone-BPE}} \\
    phone-bpe-v2-128           & 128          & 18.04 & {\scriptsize[17.78, 18.32]} \\
    phone-bpe-v2-192           & 192          & 18.96 & {\scriptsize[18.69, 19.24]} \\
    phone-bpe-v2-256           & 256          & 19.55 & {\scriptsize[19.28, 19.83]} \\
    phone-bpe-v2-512           & 512          & 18.98 & {\scriptsize[18.71, 19.25]} \\
    phone-bpe-v2-1024          & 1024         & \textbf{17.46} & {\scriptsize[17.20, 17.73]} \\
    \midrule
    \multicolumn{4}{l}{\emph{Phone-Syllable}} \\
    phone-syllable-v2-128      & 128          & 18.15 & {\scriptsize[17.88, 18.42]} \\
    phone-syllable-v2-192      & 192          & 18.31 & {\scriptsize[18.04, 18.59]} \\
    phone-syllable-v2-256      & 256          & 19.57 & {\scriptsize[19.29, 19.85]} \\
    phone-syllable-v2-512      & 512          & 18.55 & {\scriptsize[18.28, 18.82]} \\
    phone-syllable-v2-1024     & 1024         & 17.70 & {\scriptsize[17.43, 17.97]} \\
    \midrule
    \multicolumn{4}{l}{\emph{Phone-Multisyl (lite, top-100 syl UDS)}} \\
    phone-multisyl-lite-v2-512  & 512         & 18.72 & {\scriptsize[18.45, 18.99]} \\
    phone-multisyl-lite-v2-1024 & 1024        & 19.52 & {\scriptsize[19.24, 19.80]} \\
    phone-multisyl-lite-v2-2048 & 2048        & 20.04 & {\scriptsize[19.77, 20.32]} \\
    \midrule
    \multicolumn{4}{l}{\emph{Phone-Multisyl (full, 95\,\% syl coverage)}} \\
    phone-multisyl-v2-2048     & 2048         & 19.64 & {\scriptsize[19.37, 19.93]} \\
    phone-multisyl-v2-3072     & 3072         & 19.96 & {\scriptsize[19.70, 20.24]} \\
    phone-multisyl-v2-3840$^{\dagger}$ & 3840 & 40.79 & {\scriptsize[40.51, 41.08]} \\
    \bottomrule
  \end{tabular}
  \caption{Phoneme-track in-domain PER on CV+SWC test ($300\,$M).
  Bold marks the best. Spread across the $17$ convergent variants is
  $\sim$$2.5$\,pp; phone-char-raw-v2 ($43$~tokens) is within reach of
  phone-bpe-v2-$1024$ with a $24\times$ smaller vocabulary.
  $^{\dagger}$phone-multisyl-v2-$3840$ has not converged at $20\,$k
  steps and is excluded from comparisons.
  \label{tab:in-domain-per}}
\end{table}
\begin{table}[t]
  \centering
  \footnotesize
  \setlength{\tabcolsep}{3pt}
  \begin{tabular}{lrrr}
    \toprule
     & \multicolumn{2}{c}{\textbf{WER\,\%}} & \textbf{CER\,\%} \\
    \cmidrule(lr){2-3}\cmidrule(lr){4-4}
    \textbf{Tokenizer} & $1\,$B & $300\,$M & $1\,$B \\
    \midrule
    \multicolumn{4}{l}{\emph{Multilingual character inventory}} \\
    omni-vocab           & \textbf{11.65} & 15.64 & \textbf{5.08} \\
    \midrule
    \multicolumn{4}{l}{\emph{Data-driven BPE (German)}} \\
    German-BPE-512       & 14.34 & \textbf{14.93} & 5.93 \\
    German-BPE-1024      & 15.90 & 16.88 & 6.47 \\
    German-BPE-2048      & 14.64 & 16.11 & 6.25 \\
    \midrule
    \multicolumn{4}{l}{\emph{Syllable (Pyphen)}} \\
    Syllable-512         & 12.27 & 15.16 & 5.35 \\
    Syllable-1024        & 13.04 & 15.44 & 5.58 \\
    Syllable-2048        & 15.44 & 15.94 & 6.32 \\
    \midrule
    \multicolumn{4}{l}{\emph{Multisyl}} \\
    Multisyl-512         & 11.88 & 15.72 & 5.25 \\
    Multisyl-nosep-512   & 13.67 & 15.87 & 5.70 \\
    \bottomrule
  \end{tabular}
  \caption{In-domain WER/CER on CV+SWC test ($n{=}24{,}604$) for the nine
  tokenizers also fine-tuned at $1\,$B (continued-training checkpoints),
  alongside their $300\,$M WER from Table~\ref{tab:in-domain}. Scoring as
  in Table~\ref{tab:in-domain}; bold marks the best per column.
  \label{tab:in-domain-1b}}
\end{table}

Table~\ref{tab:in-domain} reports CV+SWC test WER \emph{and} CER for the $300\,$M orthographic models, scored on normalized text.
The WER range is $14.9$--$18.0$\,\% (German-BPE-512 best, multisyl-nosep-2048 worst); CIs have half-widths of $0.25$--$0.27$\,pp and overlap substantially across the top cluster: German-BPE-512, syllable-512, omni-vocab, syllable-1024, and the multi-syllable variants at vocabulary 512 are statistically indistinguishable from each other on WER.
The CER ranking is largely parallel but with a different leader: \emph{omni-vocab} reaches the lowest CER at $6.09$\,\%, a natural consequence of its character-level granularity.
The \textbf{phonological-bias penalty is zero in-domain} on both metrics: our phonologically informed tokenizers match BPE and character baselines.
The phoneme track (Table~\ref{tab:in-domain-per}) shows the same pattern in PER: the 17 convergent phone tokenizers all fall in the $17.5$--$20.0$\,\% band.

Table~\ref{tab:in-domain-1b} adds the in-domain results for the nine tokenizers that were also fine-tuned at $1\,$B, so that in-domain and cross-domain results (Table~\ref{tab:cross-domain}) can be read at the same scale.
Scaling the encoder lowers WER for every tokenizer, but not uniformly: at $300\,$M, German-BPE-512 leads and the phonologically informed tokenizers are tied with it; at $1\,$B the ordering inverts, with omni-vocab ($11.65$\,\%), multisyl-512 ($11.88$\,\%), and syllable-512 ($12.27$\,\%) ahead of every German-BPE variant (best $14.34$\,\%; syllable-512 vs.\ German-BPE-512: P2-1b-cv in Table~\ref{tab:sig}).

\subsection{Q2: Dialectal OOD (RVG\,1)}

\begin{table*}[t]
  \centering
  \footnotesize
  \setlength{\tabcolsep}{2pt}
  \begin{tabular}{lcccccccc}
    \toprule
    \multicolumn{9}{c}{\textbf{wav2vec\,2.0, $1\,$B (continued-training checkpoints)}} \\
    \midrule
     & \multicolumn{4}{c}{\textbf{RVG\,1 (dialectal spontaneous, $n{=}442$)}} & \multicolumn{4}{c}{\textbf{Verbmobil (spontaneous, German, $n{\approx}13{,}600$)}} \\
    \cmidrule(lr){2-5}\cmidrule(lr){6-9}
     & \multicolumn{2}{c}{ORT} & \multicolumn{2}{c}{TR2} & \multicolumn{2}{c}{ORT} & \multicolumn{2}{c}{TR2} \\
    \textbf{Tokenizer} & \textbf{WER} & \textbf{CER} & \textbf{WER} & \textbf{CER} & \textbf{WER} & \textbf{CER} & \textbf{WER} & \textbf{CER} \\
    \midrule
    \multicolumn{9}{l}{\emph{Multilingual character inventory}} \\
    omni-vocab              & 36.83{\scriptsize${\pm}$2.1} & \textbf{16.53}{\scriptsize${\pm}$1.3} & \textbf{45.35}{\scriptsize${\pm}$1.8} & \textbf{18.23}{\scriptsize${\pm}$1.1} & 22.16{\scriptsize${\pm}$0.3} & 14.08{\scriptsize${\pm}$0.2} & 22.68{\scriptsize${\pm}$0.3} & 14.10{\scriptsize${\pm}$0.2} \\
    \midrule
    \multicolumn{9}{l}{\emph{Data-driven BPE (German)}} \\
    German-BPE-512           & 47.12{\scriptsize${\pm}$2.3} & 27.75{\scriptsize${\pm}$1.9} & 54.75{\scriptsize${\pm}$2.0} & 28.79{\scriptsize${\pm}$1.8} & 25.47{\scriptsize${\pm}$0.3} & 16.63{\scriptsize${\pm}$0.3} & 26.40{\scriptsize${\pm}$0.3} & 16.61{\scriptsize${\pm}$0.3} \\
    German-BPE-1024          & 45.07{\scriptsize${\pm}$2.1} & 25.00{\scriptsize${\pm}$1.6} & 52.72{\scriptsize${\pm}$1.8} & 25.87{\scriptsize${\pm}$1.4} & 26.95{\scriptsize${\pm}$0.3} & 16.05{\scriptsize${\pm}$0.3} & 27.88{\scriptsize${\pm}$0.3} & 16.05{\scriptsize${\pm}$0.3} \\
    German-BPE-2048          & 42.80{\scriptsize${\pm}$2.0} & 25.82{\scriptsize${\pm}$1.7} & 51.14{\scriptsize${\pm}$1.8} & 26.79{\scriptsize${\pm}$1.5} & 24.17{\scriptsize${\pm}$0.3} & 15.40{\scriptsize${\pm}$0.3} & 25.14{\scriptsize${\pm}$0.3} & 15.38{\scriptsize${\pm}$0.3} \\
    \midrule
    \multicolumn{9}{l}{\emph{Syllable (Pyphen)}} \\
    syllable-512             & 38.63{\scriptsize${\pm}$2.3} & 21.77{\scriptsize${\pm}$1.8} & 48.24{\scriptsize${\pm}$2.0} & 23.48{\scriptsize${\pm}$1.6} & \textbf{21.27}{\scriptsize${\pm}$0.3} & 14.10{\scriptsize${\pm}$0.3} & \textbf{22.10}{\scriptsize${\pm}$0.3} & 14.12{\scriptsize${\pm}$0.3} \\
    syllable-1024            & 42.37{\scriptsize${\pm}$2.4} & 25.24{\scriptsize${\pm}$2.0} & 51.09{\scriptsize${\pm}$2.1} & 26.59{\scriptsize${\pm}$1.8} & 21.83{\scriptsize${\pm}$0.3} & \textbf{13.51}{\scriptsize${\pm}$0.3} & 22.83{\scriptsize${\pm}$0.3} & \textbf{13.52}{\scriptsize${\pm}$0.3} \\
    syllable-2048            & 44.31{\scriptsize${\pm}$2.1} & 25.10{\scriptsize${\pm}$1.7} & 52.10{\scriptsize${\pm}$1.9} & 26.02{\scriptsize${\pm}$1.5} & 25.60{\scriptsize${\pm}$0.3} & 15.78{\scriptsize${\pm}$0.2} & 26.39{\scriptsize${\pm}$0.3} & 15.69{\scriptsize${\pm}$0.2} \\
    \midrule
    \multicolumn{9}{l}{\emph{Multisyl}} \\
    \textbf{multisyl-512}    & \textbf{35.75}{\scriptsize${\pm}$2.0} & 18.11{\scriptsize${\pm}$1.4} & 45.62{\scriptsize${\pm}$1.8} & 20.02{\scriptsize${\pm}$1.2} & 22.05{\scriptsize${\pm}$0.3} & 14.93{\scriptsize${\pm}$0.3} & 22.77{\scriptsize${\pm}$0.3} & 14.89{\scriptsize${\pm}$0.3} \\
    \midrule
    \multicolumn{9}{l}{\emph{Multisyl-NoSep}} \\
    multisyl-nosep-512       & 45.14{\scriptsize${\pm}$2.4} & 24.22{\scriptsize${\pm}$1.8} & 53.02{\scriptsize${\pm}$2.1} & 25.43{\scriptsize${\pm}$1.6} & 23.94{\scriptsize${\pm}$0.3} & 14.47{\scriptsize${\pm}$0.3} & 24.73{\scriptsize${\pm}$0.3} & 14.43{\scriptsize${\pm}$0.3} \\
    \bottomrule
  \end{tabular}
  \caption{Cross-domain WER/CER (\%) at $1\,$B on normalized text
  (lowercase, ß~$\!\to\!$~ss, punctuation stripped), by tokenizer family.
  RVG\,1: pooled train+dev+test ($n{=}442$, audio chunked at silence to
  ${\le}60$\,s); Verbmobil: German subset, pooled ($n{=}13{,}613$ ORT /
  $13{,}618$ TR2 scored for all nine systems).
  $\pm$ = paired-bootstrap 95\,\% CI half-width
  ($n_\mathrm{boot}{=}10{,}000$, seed=$42$); bold = best per column.
  Significance tests: Table~\ref{tab:sig}.
  \label{tab:cross-domain}}
\end{table*}

Table~\ref{tab:cross-domain} (left half) shows the $1\,$B cross-domain results (continued-training checkpoints), scored on normalized text.
On the ORT tier, \textbf{multisyl-512-1b reaches the lowest WER at $\mathbf{35.75}$\,\%}, ahead of omni-vocab-1b at $36.83$\,\% (P1-ort in Table~\ref{tab:sig}).
On CER, omni-vocab is best at $16.53$ vs.\ multisyl-512 at $18.11$.
On the TR2 tier the two systems are statistically indistinguishable ($45.62$ vs.\ $45.35$; P1-tr2). On ORT, syllable-512-1b is third ($38.63$ / $21.77$).

This pattern is consistent with syllable-aware tokenization absorbing dialectal phonetic variation (aligned examples in Appendix~\ref{app:examples}).
Each German dialect realizes the canonical syllable slightly differently, and we hypothesize that syllable-level CTC targets encourage the encoder to focus on syllable-aligned representations that are more stable than character-level detail across phonetic variation; prior work on emergent syllable-like structure in a visually grounded self-supervised speech model \citep{peng2023syllable} is consistent with this picture.
The character-level multilingual baseline must instead reconcile every character with every variant pronunciation.
A direct test would require representation-level analysis (e.g., diagnostic classifiers for syllable boundaries), which we leave to future work.
The fact that the advantage emerges on ORT but is statistically absent on the phonetically faithful TR2 tier (P1-tr2) further suggests that the syllable advantage on RVG\,1-ORT depends on the canonical-orthography reference and weakens when the reference includes dialect markers.

\subsection{Q3: Spontaneous-lexical OOD (Verbmobil)}

Table~\ref{tab:cross-domain} (right half) shows a much flatter picture.
On the ORT tier, \textbf{syllable-512-1b reaches the lowest WER at $\mathbf{21.27}$\,\%}, ahead of multisyl-512-1b ($22.05$\,\%) and omni-vocab-1b ($22.16$\,\%), which are statistically indistinguishable from each other (C1-vm in Table~\ref{tab:sig}); syllable-1024 leads on CER ($13.51$ vs.\ $14.08$ for omni-vocab).
All German-BPE variants and syllable-2048 fall between $24.17$ and $26.95$\,\%, at least $2$\,pp behind these leaders.
The TR2 tier mirrors ORT throughout ($0.5$ to $1$\,pp higher, unchanged ordering).
The systems in front are therefore the phonologically informed tokenizers at $|V|{\le}1024$ together with the character baseline, and syllable-2048 falls $4.3$\,pp behind syllable-512, the same direction as on RVG\,1 (§\ref{sec:vocab}).

The shift on Verbmobil is driven by new lexical material rather than phonetic variation: hesitations (``\textipa{\textepsilon m}'', ``naja''), phonetic-shorthand verb conjugations (``müsten'' for ``müssten''), and discourse particles.
A character tokenizer has no out-of-vocabulary problem, whereas a syllable- or BPE-based vocabulary fails on these unseen forms via cascading substitutions (aligned examples in Appendix~\ref{app:examples}).
One consistent (though confounded with vocabulary type) interpretation is that finer-grained inventories simply have more recovery paths when no whole word matches: omni-vocab emits $\approx$\,$6.9$ tokens per word in our test data, multisyl-512 only $\approx$\,$4.5$.
In aggregate WER this advantage is offset by the character inventory's own failure mode of transcribing reduced function words literally, so on Verbmobil the two error types cancel rather than one dominating.

\begin{table*}[t]
  \centering
  \footnotesize
  \setlength{\tabcolsep}{3pt}
  \begin{tabular}{llllcr}
    \toprule
    \textbf{ID} & \textbf{Comparison} & \textbf{Setup} & \textbf{WER A / B} & \textbf{$\Delta$ (95\,\% CI)} & \textbf{$p_\mathrm{boot}$} \\
    \midrule
    P1-ort       & msy-512 vs omni @ 1B   & RVG\,1 ORT     & 35.75 / 36.83 & $-1.1$\ {\scriptsize[$-2.0,-0.2$]}    & $0.019$ \\
    P1-tr2       & msy-512 vs omni @ 1B   & RVG\,1 TR2     & 45.62 / 45.35 & $+0.3$\ {\scriptsize[$-0.5,+1.0$]}    & $0.518$ \\
    \midrule
    P2-300m-cv   & syl-512 vs gBPE-512 @ 300M & CV+SWC      & 15.16 / 14.93 & $+0.2$\ {\scriptsize[$+0.1,+0.3$]}    & $<\!0.001^{\ast}$ \\
    P2-1b-cv     & syl-512 vs gBPE-512 @ 1B   & CV+SWC      & 12.27 / 14.34 & $-2.1$\ {\scriptsize[$-2.2,-2.0$]}    & $<\!0.001^{\ast}$ \\
    P2-300m-ort  & syl-512 vs gBPE-512 @ 300M & RVG\,1 ORT      & 46.23 / 42.30 & $+3.9$\ {\scriptsize[$+3.3,+4.7$]}    & $<\!0.001^{\ast}$ \\
    P2-1b-ort    & syl-512 vs gBPE-512 @ 1B   & RVG\,1 ORT      & 38.63 / 47.12 & $-8.5$\ {\scriptsize[$-9.8,-7.2$]}    & $<\!0.001^{\ast}$ \\
    P2-300m-tr2  & syl-512 vs gBPE-512 @ 300M & RVG\,1 TR2      & 53.24 / 49.81 & $+3.4$\ {\scriptsize[$+2.8,+4.0$]}    & $<\!0.001^{\ast}$ \\
    P2-1b-tr2    & syl-512 vs gBPE-512 @ 1B   & RVG\,1 TR2      & 48.24 / 54.75 & $-6.5$\ {\scriptsize[$-7.6,-5.4$]}    & $<\!0.001^{\ast}$ \\
    \midrule
    P3           & p-char-raw vs p-BPE-1024 @ 300M & CV+SWC (PER) & 18.05 / 17.46 & $+0.6$\ {\scriptsize[$+0.5,+0.7$]}    & $<\!0.001^{\ast}$ \\
    \midrule
    \multicolumn{6}{l}{\emph{Interaction contrasts at $1\,$B (separate Bonferroni family, $k{=}5$)}} \\
    C1-vm        & msy-512 vs omni @ 1B            & Verbmobil ORT & 22.05 / 22.16   & $-0.1$\ {\scriptsize[$-0.3,+0.1$]}    & $0.214$ \\
    C2-ort       & (syl $-$ gBPE) @2048 $-$ @512   & RVG\,1 ORT    & $-8.5$ / $+1.5$ & $+10.0$\ {\scriptsize[$+8.4,+11.6$]}  & $<\!0.001^{\ast}$ \\
    C2-vm        & (syl $-$ gBPE) @2048 $-$ @512   & Verbmobil ORT & $-4.2$ / $+1.4$ & $+5.6$\ {\scriptsize[$+5.4,+5.9$]}    & $<\!0.001^{\ast}$ \\
    C3-ort       & syl-1024 vs msy-nosep-512 @ 1B  & RVG\,1 ORT    & 42.37 / 45.14   & $-2.8$\ {\scriptsize[$-3.9,-1.6$]}    & $<\!0.001^{\ast}$ \\
    C3-vm        & syl-1024 vs msy-nosep-512 @ 1B  & Verbmobil ORT & 21.83 / 23.94   & $-2.1$\ {\scriptsize[$-2.3,-1.9$]}    & $<\!0.001^{\ast}$ \\
    \bottomrule
  \end{tabular}
  \caption{Paired-bootstrap tests on $\Delta$-WER (10{,}000 resamples,
  seed=42; normalized text as in Table~\ref{tab:in-domain}). Top: nine
  focal comparisons. Bottom: interaction contrasts at $1\,$B; C1 on RVG\,1 is P1-ort, and
  for C2 the WER column lists the two constituent differences
  $\Delta_{512}$~/~$\Delta_{2048}$ (syllable minus German-BPE) whose
  difference is the reported $\Delta$. Verbmobil contrasts use the German
  subset (Table~\ref{tab:cross-domain}). $\ast$ = passes per-dataset
  Bonferroni correction ($\alpha{=}0.05/k$). $1\,$B values use the
  continued-training checkpoints, $300\,$M and phoneme-track values their
  single-stage runs. \emph{omni}~=~omni-vocab, \emph{msy}~=~multisyl,
  \emph{syl}~=~syllable, \emph{gBPE}~=~German-BPE, \emph{p-}~=~phone track.
  \label{tab:sig}}
\end{table*}

Pairwise significance tests on nine focal comparisons (Table~\ref{tab:sig}, top block) confirm this picture: every comparison on CV+SWC passes Bonferroni correction at $\alpha{=}0.05$; the syllable-vs-German-BPE comparisons on RVG\,1 are strongly significant in both directions; and the multisyl-vs-omni-vocab comparison on RVG\,1 sits at the Bonferroni boundary on the ORT tier (P1-ort) and fails significance on the TR2 tier (P1-tr2), as it does on Verbmobil (C1-vm).
For absolute scale, Whisper-large-v3 \citep{radford2023whisper} and Parakeet-TDT-v3 \citep{sekoyan2025canary}, both trained on two to three orders of magnitude more audio, reach $11.4$\,\% and $11.9$\,\% in-domain WER on the same CV+SWC test set (statistically tied with our best $1\,$B fine-tune at $11.7$\,\%) but sit $5$ to $20$\,pp ahead cross-domain.
We report these numbers for context only; training-data scale is orthogonal to the tokenizer question and is not the contribution of this work.

\subsection{Vocabulary size and tokenizer family interact}
\label{sec:vocab}

\begin{table}[t]
  \centering
  \footnotesize
  \setlength{\tabcolsep}{3pt}
  \begin{tabular}{lcc}
    \toprule
    & \textbf{RVG\,1-ORT} & \textbf{Verbmobil-ORT} \\
    $|V|$ & gBPE\,/\,Syl ($\Delta$) & gBPE\,/\,Syl ($\Delta$) \\
    \midrule
    512  & 47.12\,/\,38.63 ($-8.5$) & 25.47\,/\,21.27 ($-4.2$) \\
    1024 & 45.07\,/\,42.37 ($-2.7$) & 26.95\,/\,21.83 ($-5.1$) \\
    2048 & 42.80\,/\,44.31 ($+1.5$) & 24.17\,/\,25.60 ($+1.4$) \\
    \midrule
    $\Delta_{2048}{-}\Delta_{512}$ & $+10.0^{\ast}$ & $+5.6^{\ast}$ \\
    \bottomrule
  \end{tabular}
  \caption{WER (\%) at $1\,$B by vocabulary size for German-BPE (gBPE)
  and Syllable (Syl) on the two OOD sets (ORT tier; Verbmobil: German
  subset). $\Delta$ = Syllable
  $-$ German-BPE; the last row is the paired-bootstrap interaction
  contrast (C2 in Table~\ref{tab:sig}; $^{\ast}p<0.001$).
  \label{tab:vocab}}
\end{table}

In-domain, both effects are small: marginal WER varies by $1.0$\,pp across $|V|$ and by $1.2$\,pp across families, and $|V|{=}512$ is best or tied within every family.
Under domain shift the two factors interact (Table~\ref{tab:vocab}).
As $|V|$ grows, German-BPE improves on RVG\,1 while Syllable degrades, so the $8.5$\,pp lead of Syllable at $|V|{=}512$ turns into a $1.5$\,pp deficit at $2048$; Verbmobil shows the same crossover ($-4.2$ to $+1.4$\,pp).
The paired-bootstrap interaction contrast is $+10.0$\,pp on RVG\,1 and $+5.6$\,pp on Verbmobil (C2 in Table~\ref{tab:sig}).

Granularity alone cannot account for this pattern.
Finer tokenizers are better on average (Spearman $\rho$ between characters per token and WER across the nine $1\,$B systems: $+0.67$ on RVG\,1-ORT, $+0.68$ on Verbmobil-ORT), but Syllable is the finer tokenizer at every $|V|$, so a monotone effect of granularity predicts no crossover at all, and a granularity-matched pair still differs significantly on both corpora (C3).
Undertraining of large vocabularies does not account for it either: with the audio budget fixed, Syllable-2048 receives $1.6\times$ more target-token occurrences per vocabulary entry than German-BPE-2048 and still loses to it (Appendix~\ref{app:budget}).
We therefore read the crossover as a capacity effect: at small $|V|$ the phonological prior supplies structure that BPE cannot afford to represent, whereas at large $|V|$ BPE rediscovers syllable-like units on its own and the hard syllable constraint becomes a liability.
C3 replicates on a second, architecturally different backbone, as do the directions of C1 and C2 on RVG\,1 (mHuBERT-147; Appendix~\ref{app:mhubert}).

\subsection{Analysis: error patterns are tokenizer-invariant}

We compute confusion matrices for all $18$ phoneme tokenizers across all three corpora.
The pattern is striking: across every tokenizer the top-$5$ in-domain substitution pairs are the same bidirectional function-word swaps (``der''$\leftrightarrow$``die'', ``und''$\leftrightarrow$``der'', ``und''$\leftrightarrow$``die''), with maximum count differences between tokenizers $\leq$$10\,\%$ for the $17$ well-converged runs (the $40$\,\%-PER outlier phone-multisyl-v2-$3840$ shows the same ranking at inflated absolute counts).
This invariance holds across representative extremes of vocabulary size and merge strategy (e.g., phone-char-raw, phone-bpe-1024, phone-multisyl-lite-512).
One important caveat remains: the confusion counts are computed at the whitespace-separated SAMPA-word level rather than at the atomic-IPA-segment level (see Limitations).
Atomic-phoneme confusion analysis is left for future work; we therefore read the present observation as indirect evidence that the encoder determines \emph{which} errors occur on these data, while the tokenizer modulates \emph{how many}. In §\ref{sec:discussion} we show that this redistribution materially affects deployment WER under domain shift.

\subsection{Discussion}
\label{sec:discussion}

The opening question, whether \emph{phonologically informed} tokenizers built from German syllabification and grapheme-to-phoneme mappings might offer something that data-driven and multilingual defaults do not, admits a nuanced answer in three positive components.

\textbf{Phonologically informed tokenizers are fully competitive in-domain.}
All top tokenizers cluster within their confidence intervals on CV+SWC, so adopting syllable- or phoneme-derived units incurs no in-domain accuracy cost while adding practical advantages: more interpretable confusion analyses, smaller embedding tables in production (relative to the $10{,}288$-token multilingual character inventory), and a linguistically transparent design space.

\textbf{Under domain shift, the choice of output unit becomes a useful tool whose payoff depends on vocabulary size rather than on the type of shift.}
When the shift is phonetic (RVG\,1-ORT), the syllable-aware multisyl-512 takes the WER lead while omni-vocab stays ahead on CER; when it is lexical (Verbmobil), syllable-512 takes the lead with omni-vocab about one point behind and every German-BPE variant further back.
The item-level mechanisms still move in the direction phonological theory would predict: syllable-aligned units anchor the model in intra-syllabic acoustic variation, and finer-grained inventories absorb out-of-vocabulary lexical material. On Verbmobil, however, they offset each other in aggregate WER.
What separates the systems there is the size of the vocabulary: against German-BPE, the syllable prior wins by $4$ to $8$\,pp at $|V|{=}512$ and loses on both corpora at $|V|{=}2048$.
Tokenization thus emerges as a deployment-time lever, but one tuned against the vocabulary budget as much as against the expected mode of variation.

\textbf{The acoustic encoder and the output tokenizer act as complementary, not competing, design axes.}
Across all 18 phoneme-track tokenizers, the dominant error patterns are broadly similar bidirectional function-word swaps, indicating that tokenization redistributes errors along a shared encoder-fixed backbone, and this redistribution is large enough on domain-shifted data to materially change deployment outcomes.
We therefore see linguistic structure not as an alternative to encoder-side improvements such as domain-adaptive pretraining \citep{hsu2021robust} but as an orthogonal lever that adds value on top of them.

\section{Conclusion}
\label{sec:conclusion}

Across 40 CTC fine-tunes of the Omnilingual~ASR wav2vec\,2.0 backbone for German (31 tokenizers at $300\,$M, nine at $1\,$B), evaluated on three orthogonal corpora, the phonologically informed tokenizers (Pyphen syllables, multi-syllable merges, phoneme-derived units) are competitive with BPE and character baselines in-domain, provide the best system on both OOD sets at small vocabulary sizes while losing that lead at larger ones, and complement rather than replace the acoustic encoder, since the dominant error patterns are broadly similar across all 18 phoneme-track variants.
Linguistic structure thus remains a useful design axis for German ASR, and combining linguistic priors with encoder-side adaptation is a natural next step, as is LM rescoring, whose interaction with the output unit we leave open.

\section*{Limitations}

Our largest models are $1\,$B parameters trained on $961$\,h of audio; external baselines trained on two to three orders of magnitude more data sit $5$--$20$\,pp ahead cross-domain (although in-domain WER is statistically tied), so the trade-offs we report could be partially compressed at production scale.
We probe two structurally orthogonal OOD axes, dialectal-phonetic (RVG\,1) and spontaneous-lexical (Verbmobil), and leave combined modes (e.g., dialectal negotiation speech), noise robustness, and code-switching to future work.
All $1\,$B results in Tables~\ref{tab:in-domain-1b}, \ref{tab:cross-domain}, and~\ref{tab:sig} use the continued-training checkpoints; the in-domain delta between initial and continued training is small ($\leq$$0.7$\,pp WER across the nine $1\,$B tokenizers), so the reported cross-domain effects are not dominated by undertraining of any single variant.
The second-backbone replication (Appendix~\ref{app:mhubert}) uses a $95\,$M base model with much higher absolute error rates, and the direction of the vocabulary-size interaction C2 depends on encoder capacity.
The phoneme-track confusion analysis operates on whitespace-separated SAMPA word transcriptions rather than atomic IPA segments, and the phoneme track itself emits phoneme sequences; a deployed system would additionally require a phoneme-to-grapheme decoder \citep{lee2023twopass}.
Findings are reported for German; the dialectal-vs-spontaneous trade-off is plausible for typologically similar languages (dialectally rich, syntactically free), but cross-linguistic generalization remains an open question we leave to future work.

\section*{Ethical Considerations}

The training data (Common Voice, Spoken Wikipedia Corpus) is released under permissive licenses and consists of recordings by consenting volunteer speakers.
The OOD corpora (RVG\,1, Verbmobil) are licensed via the Bavarian Archive for Speech Signals (BAS) for research use.
The dialectal split of RVG\,1 covers a balanced regional sample of German speakers; our evaluation does not infer or release any speaker identity.
ASR systems can amplify existing inequities in recognition quality across speaker groups; our findings suggest that tokenizer choice can be tuned to mitigate this for particular OOD axes (dialectal speech in our case), and we recommend that practitioners audit their tokenizer choice for the specific deployment population.

\section*{Acknowledgments}

Funded by the Deutsche Forschungsgemeinschaft (DFG) -- Project Number 549142762 -- FIP 160.

\appendix

\section{Replication on a Second Backbone}
\label{app:mhubert}

\begin{table*}[t]
  \centering
  \footnotesize
  \setlength{\tabcolsep}{4pt}
  \begin{tabular}{llcrcr}
    \toprule
    & & \multicolumn{2}{c}{\textbf{wav2vec\,2.0, $1\,$B (Table~\ref{tab:sig})}} & \multicolumn{2}{c}{\textbf{mHuBERT-147, $95\,$M}} \\
    \cmidrule(lr){3-4}\cmidrule(lr){5-6}
    \textbf{Contrast} & \textbf{Set} & $\Delta$ (95\,\% CI) & $p_\mathrm{boot}$ & $\Delta$ (95\,\% CI) & $p_\mathrm{boot}$ \\
    \midrule
    C1: msy-512 $-$ omni              & RVG\,1 ORT    & $-1.1$\ {\scriptsize[$-2.0,-0.2$]}   & $0.019$        & $-1.0$\ {\scriptsize[$-1.7,-0.3$]}   & $0.006$ \\
                                      & Verbmobil ORT & $-0.1$\ {\scriptsize[$-0.3,+0.1$]}   & $0.214$        & $-0.3$\ {\scriptsize[$-0.5,-0.1$]}   & $<\!0.001$ \\
    \midrule
    C2: (syl $-$ gBPE)@2048 $-$ @512  & RVG\,1 ORT    & $+10.0$\ {\scriptsize[$+8.4,+11.6$]} & $<\!0.001$     & $+8.7$\ {\scriptsize[$+7.8,+9.7$]}   & $<\!0.001$ \\
                                      & Verbmobil ORT & $+5.6$\ {\scriptsize[$+5.4,+5.9$]}   & $<\!0.001$     & $-0.6$\ {\scriptsize[$-0.8,-0.4$]}   & $<\!0.001$ \\
    \midrule
    C3: syl-1024 $-$ msy-nosep-512    & RVG\,1 ORT    & $-2.8$\ {\scriptsize[$-3.9,-1.6$]}   & $<\!0.001$     & $-2.1$\ {\scriptsize[$-2.6,-1.5$]}   & $<\!0.001$ \\
                                      & Verbmobil ORT & $-2.1$\ {\scriptsize[$-2.3,-1.9$]}   & $<\!0.001$     & $-0.8$\ {\scriptsize[$-0.9,-0.6$]}   & $<\!0.001$ \\
    \bottomrule
  \end{tabular}
  \caption{Interaction contrasts C1--C3 ($\Delta$-WER in pp, paired
  bootstrap, 10{,}000 resamples, seed=42) on the two backbones; scoring and
  abbreviations as in Table~\ref{tab:sig}; Verbmobil on the German subset.
  \label{tab:mhubert-contrasts}}
\end{table*}
\begin{table}[t]
  \centering
  \footnotesize
  \setlength{\tabcolsep}{3pt}
  \begin{tabular}{lrrr}
    \toprule
    & \textbf{RVG\,1} & \textbf{Verbmobil} & \textbf{CV+SWC} \\
    \textbf{Tokenizer} & ORT & ORT & test \\
    \midrule
    omni-vocab          & 69.50 & 45.36 & \textbf{22.35} \\
    German-BPE-512      & 70.36 & 45.23 & 24.38 \\
    German-BPE-2048     & \textbf{63.27} & 46.57 & 28.24 \\
    syllable-512        & 63.32 & \textbf{43.44} & 23.26 \\
    syllable-1024       & 64.04 & 44.51 & 24.20 \\
    syllable-2048       & 64.95 & 44.18 & 25.37 \\
    multisyl-512        & 68.52 & 45.05 & 23.20 \\
    multisyl-nosep-512  & 66.11 & 45.26 & 24.09 \\
    \bottomrule
  \end{tabular}
  \caption{WER (\%) of the eight mHuBERT-147 fine-tunes ($30\,$k steps,
  $10\,$k of them with the encoder frozen); test sets and scoring as in
  Tables~\ref{tab:in-domain} and~\ref{tab:cross-domain} (Verbmobil: German
  subset). Bold marks the best per column.
  \label{tab:mhubert-systems}}
\end{table}

To test whether the interaction structure of §\ref{sec:vocab} is specific to the wav2vec\,2.0 encoder, we repeat the experiment with mHuBERT-147 \citep{boito2024mhubert}, a $95\,$M-parameter multilingual HuBERT base model that differs from Omnilingual~ASR in self-supervised objective, architecture, and pretraining mixture.
The encoder is the only change: the CTC recipe, training data, tokenizer artifacts, optimizer, learning-rate schedule, and audio budget of §\ref{sec:experiments} are reused unchanged.
Because mHuBERT-147 is a raw self-supervised checkpoint rather than an ASR-adapted one, the encoder is frozen for the first $10\,$k steps while the CTC head is trained (the standard HuBERT fine-tuning recipe; the setting is tokenizer-independent), and training runs for $30\,$k steps so that the encoder receives the same $20\,$k fine-tuning updates as in the main experiments.
We train the eight systems that enter the contrasts C1--C3 and evaluate them with the protocol of §\ref{sec:eval}.

Table~\ref{tab:mhubert-contrasts} reports the three paired-bootstrap contrasts on both backbones; Table~\ref{tab:mhubert-systems} gives the per-system WER.
Absolute error rates are two to three times higher than at $1\,$B, as expected for a $95\,$M base model; the test is whether the sign structure replicates.
C3 does so on both corpora, and C1 on RVG\,1: multisyl-512 again beats omni-vocab on dialectal speech, and the granularity-matched pair again differs significantly on RVG\,1 and on Verbmobil.
C2 replicates on RVG\,1 with a similar magnitude ($+8.7$ vs.\ $+10.0$\,pp) but is negative on Verbmobil, where the same contrast is $-0.9$\,pp on the $300\,$M wav2vec\,2.0 fine-tunes as well: its direction tracks encoder capacity rather than the encoder family.
Per-system rankings transfer poorly between backbones (Spearman $\rho$ between the mHuBERT-147 and $300\,$M wav2vec\,2.0 rankings of the eight systems: $0.00$ on RVG\,1, $+0.57$ on Verbmobil, $+0.38$ in-domain), which is what the interaction reading predicts: a ranking is not a property of the tokenizer alone.
An interaction that survives an encoder swap cannot be a property of the wav2vec\,2.0--tokenizer match.

\section{Aligned Recognition Examples}
\label{app:examples}

\begin{table*}[t]
  \centering
  \footnotesize
  \setlength{\tabcolsep}{4pt}
  \begin{tabular}{@{}p{\dimexpr0.25\textwidth-2\tabcolsep\relax}p{\dimexpr0.25\textwidth-2\tabcolsep\relax}p{\dimexpr0.25\textwidth-2\tabcolsep\relax}p{\dimexpr0.25\textwidth-2\tabcolsep\relax}@{}}
    \toprule
    \textbf{Reference} & \textbf{omni-vocab} & \textbf{multisyl-512} & \textbf{German-BPE-512} \\
    \midrule
    \multicolumn{4}{l}{\emph{RVG\,1}} \\
    und dann habe ich gedacht & und \textbf{ann} \textbf{ab} \textbf{ch} gedacht & und dann habe ich gedacht & und dann habe ich gedacht \\
    da konnte man etwas erzählen & da \textbf{konte} \textbf{an} \textbf{etas} erzählen & da konnte man etwas erzählen & da \textbf{konn} \textbf{manwas} erzählen \\
    und dann haben wir jetzt & und \textbf{dan} \textbf{aben} \textbf{ir} jetzt & und dann haben wir jetzt & und dann haben wir jetzt \\
    tag verdienen hektik gab es auch & tag \textbf{erdienen} \textbf{hekti} gab \textbf{s} auch & tag verdienen hektik gab es auch & tag \textbf{verdiener} hektik gab es auch \\
    und die alarmanlage muss man$^{\dagger}$ & und die alarmanlage muss man & und \textbf{der} \textbf{lamanlage} \textbf{mussss} man & und \textbf{der} \textbf{lamanlage} muss man \\
    \midrule
    \multicolumn{4}{l}{\emph{Verbmobil (German utterances)}} \\
    gerne herrn schiel gesprochen & erne herrn schiel gesprochen & \textbf{gerner} herrn \textbf{schkiiel} gesprochen & \textbf{undner} herrn schiel gesprochen \\
    ein anderthalbtägiges arbeitstreffen in & ein anderthalbtägiges arbeitstreffen in & ein anderthalbtägiges \textbf{arbeitsstreffen} in & ein \textbf{anderthalbägiges} arbeitstreffen in \\
    ihnen unser eineinhalbtägiges arbeitstreffen & ihnen unser eineinhalbtägiges arbeitstreffen & ihnen unser eineinhalbtägiges \textbf{arbeittreffen} & ihnen unser \textbf{eineinhalbttägiges} arbeitstreffen \\
    schindel ich möchte mit ihnen gerne einige arbeitssitzungen & schindel ich möchte mit ihnen gerne einige arbeitssitzungen & \textbf{schinl} ich möchte mit ihnen gerne einige \textbf{arbeitsetzungen} & \textbf{schinel} ich möchte mit ihnen gerne einige arbeitssitzungen \\
    dann könnten wir uns vielleicht auch noch$^{\dagger}$ & dann \textbf{können} wir \textbf{us} \textbf{vieleict} \textbf{ach} noch & dann könnten wir uns vielleicht auch noch & dann könnten wir uns vielleicht auch noch \\
    \bottomrule
  \end{tabular}
  \caption{Automatically selected reference/hypothesis windows ($1\,$B systems,
  normalized text). Bold marks hypothesis words that do not match the
  reference. Selection: RVG\,1 windows in which multisyl-512 is error-free
  and every omni-vocab error is a near-miss substitution (character distance
  ${\le}2$); Verbmobil windows in which omni-vocab is error-free and both
  subword systems err on a word absent from the training text.
  $^{\dagger}$Counterexample with the opposite pattern.
  \label{tab:examples}}
\end{table*}
\begin{table*}[t]
  \centering
  \footnotesize
  \setlength{\tabcolsep}{4pt}
  \begin{tabular}{lp{0.8\textwidth}}
    \toprule
    \textbf{System} & \textbf{Most frequent word substitutions (reference$\to$hypothesis, count)} \\
    \midrule
    \multicolumn{2}{l}{\emph{RVG\,1-ORT}} \\
    omni-vocab & habe$\to$hab (296), das$\to$es (76), es$\to$s (69), das$\to$des (61), dann$\to$dan (45), ist$\to$is (44), dass$\to$das (42), und$\to$un (35) \\
    multisyl-512 & habe$\to$ha (232), habe$\to$hab (100), das$\to$es (60), halt$\to$ha (55), das$\to$des (41), dass$\to$dasss (38), noch$\to$nur (38), dass$\to$das (36) \\
    German-BPE-512 & ja$\to$j (70), was$\to$w (50), das$\to$es (46), bisschen$\to$bis (42), habe$\to$hab (38), habe$\to$h (32), bin$\to$b (32), das$\to$des (29) \\
    \midrule
    \multicolumn{2}{l}{\emph{Verbmobil-ORT (German)}} \\
    omni-vocab & wäre$\to$wär (767), es$\to$s (637), habe$\to$hab (469), das$\to$es (327), einen$\to$ein (211), würde$\to$wird (174), ginge$\to$ging (160), mal$\to$nochmal (159) \\
    multisyl-512 & wäre$\to$wä (418), wäre$\to$wär (401), es$\to$s (330), das$\to$es (303), habe$\to$hab (299), wiederhören$\to$hören (298), mal$\to$nochmal (229), habe$\to$ha (181) \\
    German-BPE-512 & wäre$\to$wä (466), das$\to$es (343), einen$\to$ein (293), ja$\to$j (246), wäre$\to$w (225), habe$\to$hab (214), ist$\to$es (213), ihnen$\to$ihn (190) \\
    \bottomrule
  \end{tabular}
  \caption{Eight most frequent word substitutions per system ($1\,$B, normalized
  text) on the two OOD sets. On RVG\,1 the substitutions are dominated by
  reduced dialectal realizations of function words; on Verbmobil by colloquial
  contractions shared by all systems.
  \label{tab:subs}}
\end{table*}

Tables~\ref{tab:examples} and~\ref{tab:subs} illustrate the error types behind the cross-domain results, using the $1\,$B systems omni-vocab, multisyl-512, and German-BPE-512 (one per tokenizer class).
Examples are selected automatically rather than by hand: we align reference and hypothesis at the word level, slide windows of 4--10 reference words over every utterance, and discard windows in which any system's hypothesis collapses (hypothesis length outside 0.6--1.5 times the window length).
On RVG\,1 we keep windows in which multisyl-512 is error-free while every omni-vocab error is a near-miss substitution (character distance ${\le}2$), i.e., a phonetic rendering of a dialectal realization; on Verbmobil (German utterances) we keep windows in which omni-vocab is error-free while both subword systems err on a word absent from the $6.14$\,M-word training text.
Windows are ranked by the number of such errors, at most one window per utterance; one counterexample with the opposite pattern is shown per corpus.

The two corpora show different error types.
On RVG\,1 the character inventory transcribes reduced dialectal forms literally (\emph{habe}$\to$\emph{hab}, \emph{ist}$\to$\emph{is}, \emph{und}$\to$\emph{un}) and drops consonants inside function words, whereas syllable-constrained units restore the canonical form.
On Verbmobil the subword systems fail on out-of-vocabulary compounds and names by producing plausible but wrong forms (\emph{arbeitsstreffen}, \emph{anderthalbägiges}), while the character inventory spells them correctly; the most frequent substitutions (Table~\ref{tab:subs}), colloquial contractions such as \emph{wäre}$\to$\emph{wär} and \emph{es}$\to$\emph{s}, are shared by all systems.

\section{Target-Token Statistics per Run}
\label{app:budget}

\begin{table}[t]
  \centering
  \footnotesize
  \setlength{\tabcolsep}{3.5pt}
  \begin{tabular}{lrrrrr}
    \toprule
    \textbf{Tokenizer} & $|V|$ & \textbf{tok/wd} & \textbf{ch/tok} & \textbf{tokens} & \textbf{per} \\
                       &       &                 &                 & \textbf{(M)}    & \textbf{entry (k)} \\
    \midrule
    \multicolumn{6}{l}{\emph{Multilingual character inventory}} \\
    omni-vocab & 10{,}288 & 6.96 & 0.87 & 42.7 & 4.2 \\
    \midrule
    \multicolumn{6}{l}{\emph{Data-driven BPE (German)}} \\
    German-BPE-512 & 512 & 2.72 & 2.23 & 16.7 & 32.6 \\
    German-BPE-1024 & 1024 & 2.32 & 2.61 & 14.2 & 13.9 \\
    German-BPE-2048 & 2048 & 1.99 & 3.04 & 12.2 & 6.0 \\
    \midrule
    \multicolumn{6}{l}{\emph{Syllable (Pyphen)}} \\
    Syllable-512 & 512 & 3.83 & 1.58 & 23.5 & 45.9 \\
    Syllable-1024 & 1024 & 3.49 & 1.74 & 21.4 & 20.9 \\
    Syllable-2048 & 2048 & 3.24 & 1.87 & 19.9 & 9.7 \\
    \midrule
    \multicolumn{6}{l}{\emph{Multisyl}} \\
    Multisyl-512 & 512 & 4.55 & 1.33 & 27.9 & 54.6 \\
    Multisyl-1024 & 1024 & 4.41 & 1.37 & 27.1 & 26.4 \\
    Multisyl-2048 & 2048 & 4.34 & 1.40 & 26.7 & 13.0 \\
    \midrule
    \multicolumn{6}{l}{\emph{Multisyl-NoSep}} \\
    Multisyl-nosep-512 & 512 & 3.51 & 1.72 & 21.6 & 42.1 \\
    Multisyl-nosep-1024 & 1024 & 3.36 & 1.80 & 20.6 & 20.2 \\
    Multisyl-nosep-2048 & 2048 & 3.26 & 1.85 & 20.0 & 9.8 \\
    \bottomrule
  \end{tabular}
  \caption{Target-token statistics for the 13 orthographic tokenizers,
  from corpus-level fertility on the training text ($550{,}747$ sentences,
  $6{,}141{,}326$ words). \emph{tokens} = tokens per word $\times$ words;
  \emph{per entry} = tokens\,/\,$|V|$, i.e., how often an average vocabulary
  entry occurs in one pass over the training text; this is distinct from gradient
  steps, which are fixed at $20\,$k for every run (§\ref{sec:experiments}). The phoneme
  track is tokenized on its own SAMPA transcripts and is not included.
  \label{tab:budget}}
\end{table}

Batching is audio-element-based (§\ref{sec:experiments}), so every run sees the same audio for the same number of updates while the number of target tokens varies with tokenizer fertility.
Table~\ref{tab:budget} reports, for each orthographic tokenizer, the corpus-level fertility on the training text, the resulting number of target tokens per pass, and the average number of token occurrences per vocabulary entry.
The last column is the quantity an undertraining account of the vocabulary-size effect (§\ref{sec:vocab}) would have to invoke, and it does not support one: at $|V|{=}2048$ the Syllable tokenizer receives $1.6\times$ more occurrences per entry than German-BPE ($9.7$\,k vs.\ $6.0$\,k) and nonetheless loses to it.

\end{document}